\documentclass[conference]{IEEEtran}

  \usepackage{graphicx}
\usepackage{pdfpages}

\usepackage[utf8]{inputenc}
\makeatletter%
\@ifclassloaded{acmart}{
  \makeatletter
  \def\mdseries@tt{m}
  \makeatother
}{}
\makeatother

\usepackage{amsthm}
\usepackage{thmtools}

\usepackage{epsfig}
\usepackage{wrapfig}
\usepackage{pgfplotstable}
\usepackage{pgfplots}
\usepgfplotslibrary{groupplots}
\pgfplotsset{compat=newest} %
\usepgfplotslibrary{groupplots}
\usepgfplotslibrary{dateplot}
\usepackage{mathrsfs,amsmath,mathtools}
\usepackage{amssymb}
\usepackage{relsize}
\usepackage{multicol}
\usepackage{bbm}

\usepackage{float}
\usepackage{hyperref}
\usepackage{microtype}
\usepackage{tikz}
\usetikzlibrary{decorations.text,calc,shapes,arrows,arrows.meta, positioning,shapes.misc,decorations.markings,decorations.markings,decorations.pathreplacing,matrix,spy}
\usepackage{pgfplots}
\usetikzlibrary{spy,calc}
\usepackage{rotating}

\usepackage{booktabs}
\usepackage{multirow}
\usepackage{adjustbox}
\usepackage{verbatim}
\usepackage[T1]{fontenc}
\usepackage{graphicx}
\usepackage{caption}
\usepackage{subcaption}
\usepackage[noend]{algpseudocode}
\usepackage{algorithm}

\usepackage{siunitx}
\usepackage{nicefrac}
\usepackage[colorinlistoftodos]{todonotes}
\usepackage{spverbatim}
\usepackage{seqsplit}

\usepackage{enumitem}
\setlist{nosep} %

\usepackage{array}
\newcolumntype{R}[2]{%
    >{\adjustbox{angle=#1,lap=\width-(#2)}\bgroup}%
    l%
    <{\egroup}%
}

\usepackage[style=ieee,firstinits=true,doi=false,sortcites=true]{biblatex}

\AtEveryBibitem{%
  \ifentrytype{misc}{%
  }{%
    \ifentrytype{unpublished}{%
    }{%
      \clearfield{url}%
      \clearfield{urldate}%
    }%
  }%
  \clearfield{issn}%
  \clearfield{isbn}%
}

\usepackage[autostyle, english=american]{csquotes}
\MakeOuterQuote{"}

\makeatletter
\pgfplotsset{
    every axis x label/.append style={
        alias=current axis xlabel
    },
    legend pos/outer south/.style={
        /pgfplots/legend style={
            at={%
                (%
                \@ifundefined{pgf@sh@ns@current axis xlabel}%
                {xticklabel cs:0.5}%
                {current axis xlabel.south}%
                )%
            },
            anchor=north
        }
    }
}
\makeatother

\pgfplotscreateplotcyclelist{MyCyclelist}{%
  {Bittersweet, mark = *, dashed},
  {Blue, mark = square*, dashed},
  {Cyan, mark = diamond*, dashed},
  {DarkOrchid, mark = triangle*, dashed},
  {Green, mark = *, dashed},
  {Magenta, mark = square*, dashed},
  {Red, mark = diamond*, dashed},
  {Gray, mark = triangle*, dashed},
  {Black, mark = *, dashed},
    {Bittersweet, mark = square*, densely dotted},
  {Blue, mark = diamond*, densely dotted},
  {Cyan, mark = triangle*, densely dotted},
  {DarkOrchid, mark = *, densely dotted},
  {Green, mark = square*, densely dotted},
  {Magenta, mark = diamond*, densely dotted},
  {Red, mark = triangle*, densely dotted},
  {Gray, mark = *, densely dotted},
  {Black, mark = square*, densely dotted},
}

\newcolumntype{t}{>{\ttfamily}l}
\newcolumntype{T}{>{\ttfamily}c}

\newcolumntype{$}{>{\global\let\currentrowstyle\relax}}
\newcolumntype{^}{>{\currentrowstyle}}

\hypersetup{draft}

\begin{document}

\title{Getting Passive Aggressive About False Positives: Patching Deployed Malware Detectors}

\author{\IEEEauthorblockN{Edward Raff}
\IEEEauthorblockA{\textit{Laboratory for Physical Science} \\
edraff@lps.umd.edu}
\IEEEauthorblockA{\textit{Booz Allen Hamilton} \\
Raff\_Edward@bah.com}
\and
\IEEEauthorblockN{Bobby Filar}
\IEEEauthorblockA{\textit{Elastic} \\
filar@elastic.co}
\and
\IEEEauthorblockN{James Holt}
\IEEEauthorblockA{\textit{Laboratory for Physical Sciences} \\
holt@lps.umd.edu}
}

\maketitle

\begin{abstract}
False positives (FPs) have been an issue of extreme importance for anti-virus (AV) systems for decades. As more security vendors turn to machine learning, alert deluge has hit critical mass with over 20\% of all alerts resulting in FPs and, in some organizations, the number reaches half of all alerts \cite{NISC2020}. This increase has resulted in fatigue, frustration, and, worst of all, neglect from security workers on SOC teams. A foundational cause for FPs is that vendors must build one global system to try and satisfy all customers, but have no method to adjust to individual local environments. This leads to outrageous, albeit technically correct, characterization of their platforms being 99.9\% effective. Once these systems are deployed the idiosyncrasies of individual, local environments expose blind spots that lead to FPs and uncertainty.

We propose a strategy for fixing false positives in production after a model has already been deployed. For too long the industry has tried to combat these problems with inefficient, and at times, dangerous allowlist techniques and excessive model retraining which is no longer enough. We propose using a technique called passive-aggressive learning to alter a malware detection model to an individual’s environment, eliminating false positives without sharing any customer sensitive information. We will show how to use passive-aggressive learning to solve a collection of notoriously difficult false positives from a production environment without compromising the malware model’s accuracy, reducing the total number of FP alerts by an average of 23x.

\end{abstract}

\section{Introduction}
Employing machine learning to detect good/bad activity in unknown environments will yield false positives (FPs). Security vendors are quick to point out the latest ML-backed feature is 99.9\% effective, and they are not wrong. But, these are models trained on a global representation of data. Once operational in a customer’s local environment, it is not uncommon for FPs to pile up. This stems from the fact that while the model may be 99.9\% accurate \textit{globally}, the customers \textit{local} environment may have very different characteristics or peculiarities. The global vs. local dichotomy is not a new observation. Previous studies have shown that other AV products can have different FP and false negative (FN) rates for regional malware (e.g., United States vs. Brazil) and malware types (e.g., Trojan vs. ransomware) ~\cite{Botacin2020}.  This can cause excessive false alerts, which users find unacceptable and may lead them to abandon the malware detectors. 

While Windows PE (Portable Executable) and Android APK (Android Package) malware detection are more common in the literature, the same is true for all domains. Our domain of interest, detecting malicious Microsoft Office Macros, suffers these same issues with additional difficulties. 

The standard approach to false positive triage, allowlisting on file hash, works well for executable applications that update infrequently but is mostly a futile approach for macro detection. This is due to the hash of the Office document changing with each modification to the file, a common occurrence. Security workers are essentially left to play "whack-a-mole," hoping that the one-to-one hash-to-file allowlisting effectively suppresses future FPs and lacks the tools/support needed to perform anything more effective ~\cite{10.1145/3319535.3354239}. Macro malware is also less prevalent, resulting in less training data, making frequent global retraining ineffective in fixing local bursts of false positives. The other commonly proposed option is to have users report all false positives back to the company and wait for a new model update to get the previous false positives correct after training. This can takes weeks to months and does not alleviate immediate concerns. This may also be untenable for users with privacy concerns who are unwilling to share the files causing false positives. 

Despite the large focus placed on having low false positives, we are aware of no prior work that has considered fixing or correcting a model's false positives once already in production. 
Our contributions are: 1) framing the need to study \textit{correcting} a deployed model in a safe (does not diverge) and effective (still obtains meaningful true-positive rates after alteration) way that is more general than allowlists. 2) An initial solution to this problem. 
We propose using a particular type of machine learning, \textit{Passive Aggressive} (PA) learning, as an approach to 
correct false positives made by a model.
Following current production deployment, analysts/users 
 mark the false positives’ files, and we perform a local adaption of their model to correct these errors. This ensures the false positives are fixed immediately. To prevent users from erroneously "correcting" on mislabeled data, we introduce a method to estimate the impact to a model's Area Under the Curve (AUC) to detect such situations, inform the user or perform some other intervention. Combined, we show that we can perform these updates while maintaining acceptable performance, fixing the false positives and reducing alerts, and detect potentially erroneous updates. 

We will review the related work to our own in \autoref{sec:related_work}, and the production dataset used for our experiments in \autoref{sec:data}. A review of the Passive Aggressive approach, and how we use it counter to the typical practice, is given in \autoref{sec:methods}. This includes our approach to estimate the impact of a change to the model in \autoref{sec:kmeans_impact}. To show the importance of using the PA approach specifically, we will detail several baseline models we compare against in \autoref{sec:models}, followed by our results showing the utility of our PA based local refinement in \autoref{sec:results}. Given our results, some practical considerations will be reviewed in \autoref{sec:practical} followed by our conclusions in \autoref{sec:conclusion}.

\section{Related Work} \label{sec:related_work}

The focus on false positives within malware detection is not new. Many prior works of varying different approaches, from static features, to dynamic analysis and even contextual information, have been built with an emphasis on achieving low false positives in the design phase (e.g., \cite{Li2017,Dahl2013a,Shafiq2009,10.5555/2028067.2028070,Fukushima2010}). These works acknowledge and design around unique costs in keeping false positives low, but we know no prior work that focuses on how to deal with false positives once a model is already in production. The current solution for most corporate AV providers is to receive the false positive reports from customers, encourage allowlist use, and provide a model update once a quarter that \textit{hopefully} resolves any troublesome false positives. This focus on false positive resolution only during the design phase of a model remains true for related areas like spam detection \cite{learning-at-low-false-positive-rates}, clustering malware families\cite{Jang2011}, and code duplication detection within malware\cite{Calleja2019}. 
For non-machine learning-based solutions, one would increase the size/specificity of signatures to reduce FP \cite{Christodorescu:2004:TMD:1007512.1007518} or in ML-based approaches, remove processing steps/features that lead to FPs \cite{182795}. Once our model is deployed, these are no longer options. It is also necessary that our model adapt quickly with the minimum possible number of false positives, as customers will not wait or tolerate hundreds or thousands of FPs to alter a model. Our approach to alter the model using the Passive-Aggressive algorithm gives us the ability to tackle this issue in a manner that recognizes customer concerns and privacy, without having to wait weeks for an updated model.

The general theme of our approach is that we need to learn using a minimal number of samples. Typically this would fall into the domain of "one-shot" or "few-shot" learning, where a model must be adapted to a new concept with a limited number of examples. Classically few-shot learning is meant for multi-class problems where new classes may be introduced at test time, and limited prior work has investigated few-shot learning for malware family detection \cite{Tran2018,Hsiao2019}. In our case, we have only two classes, and our goal is to refine the decision surface in a small manner to fix a problematic set of instances. We compare against few-shot learning as an alternative approach to our own but find that the differences in scenarios lead to poor model performance. 

The original domain knowledge-based model used XGBoost \cite{xgboost} as the learning algorithm, a popular decision tree boosting library. Since the GBDT had the best global performance, we would prefer to test modifying the GBDT to fix the false positives found in production. While prior work like Hodeffing Trees \cite{Bifet2010,Domingos2000b}, VFDT trees \cite{Hulten2001}, and Mondrian Forests \cite{NIPS2014_5234} allow online learning of decision trees, none of them support the kind of "correction" using just one or two samples that we need. Instead, these approaches require multiple samples before altering the tree structure or adjusting the prediction in a leaf node. In practice, we have too few examples of the hard false positives to alter the tree's predictions. As such, we do not consider these approaches in any detail.

It is important to note that adversarial attacks \cite{Biggio2017} are an important problem, of particular relevance to malware detection. Malware involves a real live adversary who wishes to avoid detection \cite{Rajab2011}, which makes the motivation for attacking and defending malware detection models salient and well-motivated. 
Indeed the MalConv based model we use as our foundation has had numerous examples of adversarial attacks performed against it \cite{Suciu2018a,Demetrio2019,Kolosnjaji2018,Kreuk2018}.
At the same time, there have also been approaches that counter these attacks at an impact to accuracy, by exploiting the one-way nature of attacks in this space (that \textit{only } malware authors only wish to fool the detector) \cite{Demontis2017,Incer:2018:ARM:3180445.3180449,Fleshman2018a}
Indeed, there are various pros and cons to a machine learning-based approach to detect like our own, and more classical methods, in terms of vulnerability to attack \cite{Fleshman2018}.
For this work, we consider the adversarial problem important but out-of-scope. This is motivated in part by the fact that in real-world use, more straightforward methods than adversarial attacks currently satisfy the malicious actor's desires and are most prevalent  \cite{Anderson2017a,Aghakhani2020}.

\section{Data Used and Motivation} \label{sec:data}

We crafted a problem that was motivated by real issues in deploying information security models in production environments. We base our testing and results on a representative sample of industry data. Our corpus consisted of 1,101,407 Microsoft Office documents that contained macros. Similar to \cite{rudd2018meade} data was collected from Common Crawl (CC) \cite{crawl2018common} and VirusTotal \cite{Virustotal}. All CC samples were checked against VirusTotal to ensure identical label logic was applied across all samples. If 5+ Vendors viewed the sample as malicious, we assigned a malicious label. If no vendors labeled the sample as malicious, and the sample had been submitted at least one month before our collection date, the sample was labeled benign. This approach is supported by a recent year-long study of AV labeling that recommends using a threshold $\geq 2$, and that when using all available engines, the majority of file labels are stable as AV products update their databases \cite{251586}.  This label logic yielded 651,872 benign and 449,535 malicious samples. This means the random guessing performance would be 59.2\%, the percentage of benign files in the corpus. Will use a stratified sample of 80\% for the training set, and 20\% for the test set. We will refer to this test set as the "large" test set. 

A second test set will be composed of 58 hard false positives. These files come predominantly from two larger commercial financial tools, widely used by many businesses, and considered critical to business operations\footnote{The exact origin is regarded as sensitive information, and so unfortunately not disclosed.}. These 58 files were discovered during the initial deployment of a production macro malware detector and were regularly miss-classified as malicious during multiple model iterations. These files were objectively difficult due to their large and complex macro codebases, exhibiting many of the same behaviors as other malicious macros, but were entirely benign. Because these files were considered mission-critical by customers, their failure to be classified correctly was a significant issue, and motivate this work. We will refer to these 58 files as the "hard FPs" or the hard benign set. 

From a machine learning context, most models are built with a target false positive rate (FPR) (e.g., 0.1\%), and the threshold on a model's output is adjusted to meet this desired FPR. Normally one would continue to adjust this threshold if too many FPs were occurring. We make explicit that this is not a possible solution for these 58 difficult samples. The output probability for all samples ranges from near 0\% to near 100\%. As such, adjusting the threshold would require calling almost every file benign to avoid these FPs, which is unacceptable. This situation occurred with all machine learning approaches attempted in this work.

\begin{figure*}[!htb]
    \centering
    \begin{tikzpicture}[spy using outlines={rectangle, magnification=4,connect spies}]
        \node[anchor=south west,inner sep=0] (image) at (0,0) {\includegraphics[width=0.9\textwidth]{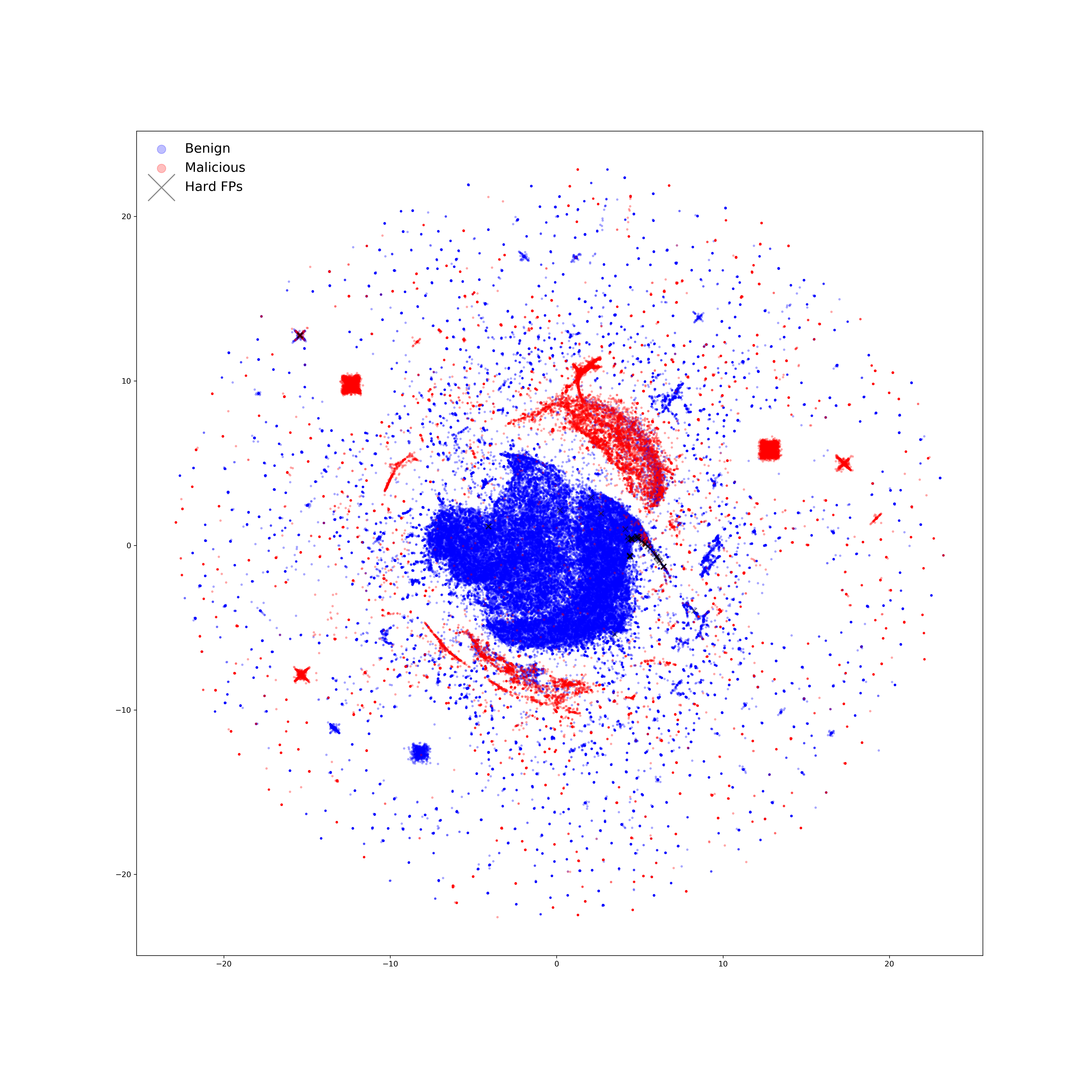}};
        \begin{scope}[x={(image.south east)},y={(image.north west)}]
            \coordinate (spypoint) at (0.60,0.48) ;
            \coordinate (spyviewer) at (0.2,0.2);
            \spy[width=4cm,height=4cm] on (spypoint) in node [fill=white] at (spyviewer);
        \end{scope}
    \end{tikzpicture}
    \caption{UMAP embedding of the Macro dataset using the penultimate activations of MalConv. Benign samples in {\color{blue} blue}, malicious samples in {\color{red} red}, and a black X marks the locations for the 58 hard FPs.The box in the lower-left shows the zoomed in view of the area that contains the majority of hard FPs.  }
    \label{fig:macro_umap}
\end{figure*}

\section{Passive Aggressive Refinement} \label{sec:methods}

Our primary approach leverages the \textit{Passive Aggressive} (PA) algorithm \cite{Crammer2006}. Under normal circumstance, the PA algorithm works in a manner similar to the classical perceptron. At the $t$'th optimization step we have a hyperplane $\mathbf{w}_t \in \mathbb{R}^{d}$, a data point $\mathbf{x}_t \in \mathbb{R}^{d}$, and a label $y_t \in \{-1, 1\}$, the goal is that $\mathbf{w}_t^\intercal \mathbf{x}_t \cdot y_t \geq 0$, indicating we predicted $\mathbf{x}_t$ to belong to class $y_t$. If we are successful in doing this, we make no change ($\mathbf{w}_{t+1} = \mathbf{w}_t$). If not, we alter it by $\mathbf{w}_{t+1} = \mathbf{w}_t + \tau_t y_t \mathbf{x}_t$, where $\tau_t$ is a constant based on the learning algorithm. In the classical perception, $\tau_t = 1$. 

When performing Passive Aggressive updates \cite{Crammer2006}, our goal is to solve the optimization problem given by \autoref{eq:pa_prob}. 
\begin{equation}
\begin{gathered}\label{eq:pa_prob}
\mathbf{w}_{t+1}=\underset{\mathbf{w} \in \mathbb{R}^{d}}{\operatorname{argmin}} \frac{1}{2}\left\|\mathbf{w}-\mathbf{w}_{t}\right\|_2^{2} \\ \text{s.t.} \quad 
\max\left(0,1-y_t\cdot \mathbf{w}^\intercal  \mathbf{x} \right)=0
\end{gathered}
\end{equation}

We want to find a new hyperplane $\mathbf{w}_{t+1}$ that is the smallest possible distance from the current solution ($\mathbf{w}_t$) as possible, while simultaneously  making the hinge loss $\ell(s)=\max(0,1-s)$ for the current item correct. This fixes the error (or "fully corrects" the hyperplane) while minimizing the change to the model. This problem has an exact solution \cite{Crammer2006}, which is given by \autoref{eq:pa_solution}. 

\begin{equation} \label{eq:pa_solution}
\mathbf{w}_{t+1}=\mathbf{w}_{t}+\tau_{t} y_{t} \mathbf{x}_{t} \text { where } \tau_{t}=\frac{1-y_t\cdot \mathbf{w}_t^\intercal  \mathbf{x}}{\left\|\mathbf{x}_{t}\right\|^{2}}
\end{equation}

Under normal circumstances, we would not actually want to use \autoref{eq:pa_solution}, because any mislabeled, noisy, or otherwise outlier datum could destroy the accuracy of the model. Indeed \textcite{Crammer2006} modified this equation to include a regularization term $C$, to limit the change to the model (e.g., $\tau_t = \min(C, \frac{1-y_t\cdot \mathbf{w}_t^\intercal  \mathbf{x}}{\left\|\mathbf{x}_{t}\right\|^{2}}$) so that it did not fully correct the label, which resulted in significantly improved accuracy. 

In our case, we will not use the PA approach to train our model, but instead to refine an already trained model that has been deployed into production. First, we will train MalConv style network \cite{MalConv} on the input features. This model would be deployed, with a threshold selected that satisfies a desired false-positive rate (e.g., 1\% or 0.1\%). With this MalConv model, we remove the softmax layer and use the penultimate activation as a featurized version of the input file. 

Users will submit their false positives to the model, which \textit{must} fully correct the error so that it is no longer a false positive using \autoref{eq:pa_solution}. We hypothesize that most false positives live near the hyperplane border, so these corrections could be made without impairing the model's performance on the global population. We use the PA updates because they induce the smallest possible change to the model to minimize the potential impact.

To support this intuition, we used Uniform Manifold Approximation and Projection (UMAP) \cite{McInnes2018,Nolet2020} to create a 2D visualization of the benign, malicious, and hard false positives discussed in \autoref{sec:data}. The resulting UMAP embedding can be found in \autoref{fig:macro_umap}, where the black X marks show the hard false positives from the evaluation set. You can see that they exist on the edge of a larger cluster, in a region that contains a broad mix of both benign and malicious samples. 

\subsection{Preventing Catastrophic Changes to Models} \label{sec:kmeans_impact}

As was mentioned above, our approach using the complete update of \autoref{eq:pa_solution} can be dangerous if a mislabeled data point is given. For a comprehensive solution that can be used in deployment, we need to warn users from performing updates that may hinder the performance of the model as a whole. However, we can not give users access to all training data, which could cause privacy issues. We would necessitate more significant compute resources to check the impact against all prior data.

\begin{algorithm}[!ht]
\caption{Estimate Impact to AUC}
\label{algo:estimate_impact}
\begin{algorithmic}[1]
\Function{EstimateAUC}{$\mathbf{w}$, $\mathbf{c}_1, \ldots, \mathbf{c}_K$,  $s(\cdot)$, and $l(\cdot)$}
    \State $\alpha \gets 0$
    \For{$i \in [1, K]$}
        \State $\hat{y} \gets \mathbf{w}^\intercal \mathbf{c_i}$
        \If{$\hat{y} \geq 0$}
            \State $\alpha \gets \alpha + s(c_i) \cdot l(c_i)$
        \Else
            \State $\alpha \gets \alpha + s(c_i) \cdot (1 - l(c_i))$
        \EndIf
    \EndFor
    \State \Return $\frac{\alpha}{\sum_{i=1}^K s(c_i)}$ 
\EndFunction
\Require Desired number of clusters $K$, MalConv embedded data points $X$
\State $\mathbf{c}_1, \ldots, \mathbf{c}_K \gets$ $K$ means computed by $K$-Means clustering of training data $X$
\State Let $s(\mathbf{c}_j)$ indicate the number data points assigned to cluster $j$
\State Let $l(\mathbf{c}_j)$ indicate the fraction of malicious items in cluster $j$ \Comment{Users get access only to $\mathbf{c}_1, \ldots, \mathbf{c}_K$, $s(\cdot)$, and $l(\cdot)$}
\State Receive new file $f$ with label $y$, that needs to be corrected. 
\State $\mathbf{x} \gets MalConv(f)$ \Comment{Extract penultimate activation from MalConv}
\State $\mathbf{\hat{w}} \gets \frac{1-y\cdot \mathbf{w}^\intercal  \mathbf{x}}{\left\|\mathbf{x}\right\|^{2}} \cdot y \cdot \mathbf{x}$ \Comment{\autoref{eq:pa_solution}}
\State $init \gets \Call{EstimateAUC}{\mathbf{w}, \mathbf{c}_1, \ldots, \mathbf{c}_K, s(\cdot), l(\cdot)}$
\State $result \gets \Call{EstimateAUC}{\mathbf{\hat{w}}, \mathbf{c}_1, \ldots, \mathbf{c}_K, s(\cdot), l(\cdot)}$
\State \Return  estimated AUC impact $result - init$
\end{algorithmic}
\end{algorithm}

We note that we can approximate the impact of our model using the $K$-Means algorithm. Clusters from the training data can be shared with users in the deployed model without risking privacy concerns and keeping the corpus small enough that evaluations can be run quickly with standard desktop hardware. 

Our approach is outlined in \autoref{algo:estimate_impact}, where the \textit{EstimateAUC} function takes in a hyperplane $\mathbf{w}$, the $K$-means computed from the training data $\mathbf{c}_1, \ldots, \mathbf{c}_K$, and two statistics: $s(\mathbf{c}_i) \in [0, N]$, the number of data points in cluster $\mathbf{c}_i$, and $l(\mathbf{c}_i) \in [0, 1]$, the fraction of positive instance (i.e., malware) in cluster $\mathbf{c}_i$. This function looks at the label assigned to each cluster based on the given hyperplane $\mathbf{w}$, and tabulates an error based on the average label of that cluster. Because the cluster center acts as a summary of many points, it is effectively similar to estimating AUC in buckets of probability score (or distance to the margin), assuming each item within a cluster would have received a similar relative classification score. 

Using this function, we can simply receive a new file $f$ from the user with a correct label $y$, that they wish to update the model with. We take the penultimate activations from MalConv as our feature vector, and compute a new weight vector $\mathbf{\hat{w}}$ using \autoref{eq:pa_solution}. We can then estimate the AUC using the original and updated solutions ($\mathbf{w}$ and $\mathbf{\hat{w}}$ respectively), and take the difference. If the result is positive, we will anticipate that the update improved the model quality. If negative, we have degraded the quality. In deployment, we may select a minimum threshold of allowed degradation (e.g., $0.05$) before preventing the user from performing any more updates, out of concern that they would negate the model's usefulness.

\section{Models Used} \label{sec:models}

Now that we have described our approach to using, against normal consideration, the exact Passive Aggressive updates, we will define the final model we use plus additional baselines. We conducted our experiments using two different sets of features, 1) the set of 328 features extracted using domain knowledge and selected by analysts, and 2) the penultimate activations from using a MalConv network \cite{MalConv} trained from the raw bytes of the Macro code. 

For the domain knowledge features, we will train a gradient boosted decision tree (GBDT) as our primary baseline, as this was the original model deployed in production. We find that using these domain knowledge-based features, the expressive power of a GBDT is necessary to achieve reasonable accuracy. To show this, we will train the PA model on this data as well, including a kernelized version using the Random Fourier Features approximation \cite{Rahimi2007} of the Radial Basis Function kernel, $K(x,x^\prime)=\exp \left(-\gamma \cdot \left\|\mathbf{x}-\mathbf{x}^{\prime}\right\|_2^{2}\right)$. These will be referred to as just \textit{GBDT}, \textit{PA}, and \textit{Kernel PA}, respectively. For PA and Kernel PA, we first unit normalized each feature. For Kernel PA, the $\gamma$ parameter was selected by grid search from  $[10^{-3}, 10^{-2}, 10^{-1}, 1, 10^1, 10^2, 10^3]$. 

The PA and Kernel PA baselines will show the importance of using the MalConv based feature extraction, which provides the benefit of not requiring analyst time to produce new features. We trained MalConv with a batch size of 64, and embedding dimension of 8, 128 filters with a filter size of 128, and a stride of 2. We specifically use a stride of 2 because the raw Macro code was encoded as UTF-8 due to non-ASCII characters, so a stride of 2 corresponds to processing the string one Unicode character at a time (though each Unicode token is broken up into two tokens embedded separately). MalConv was trained for 5 epochs, with a resulting AUC of 99.4\% and an accuracy of 48.27 on the hard FPs, so MalConv alone is not sufficient. 

For our PA model trained on top of the MalConv penultimate activation's, we denote it as \textit{MalConv+PA}. This is our primary approach, showing the utility of the full Passive Aggressive updates. To show why this full correction is needed, we will also compare with adjusting the final activation of MalConv using Stochastic Gradient Descent (SGD), which is the standard approach for gradient-based learning. This \textit{MalConv+SGD} baseline allows for online learning, but can not adapt in the small number of examples we desire to fix false positives in production. 

Our last MalConv based ablation model will use Prototype-Networks \cite{NIPS2017_6996}, a recent method for few-shot learning. It works by using a neural network $f(\cdot)$ (in our case, MalConv) to embed an input $\boldsymbol{x} \in \mathbb{R}^d$ into a new feature space $\boldsymbol{z} \in \mathbb{R}^{d'}$. The approach of the Prototype-network is to then average the embedding $\boldsymbol{z}_i$ within a class, and use the distance to the nearest embedding as the mechanism for classification. We train in the style described by \textcite{NIPS2017_6996} using MalConv as the network architecture with an output embedding $d' = 128$. Our results below will show that the large differences between our scenario with two classes instead of many results in the few-shot learning approach having limited utility.

\section{Results} \label{sec:results}

Now that we have described our approach MalConv+PA, and the five baselines we will compare against, we will show the results in two scenarios. First is the \textit{fixed} scenario, where we look at each method’s performance on the larger test set. This gives us an understanding of each model's global performance. Models that do not have a sufficiently adequate global performance are not viable for our use under any circumstances. 

The second scenario is the \textit{adaptive} one. Here we have an initial global model, which will be exposed to a sequence of hard false positives that are mission-critical and thus need to be corrected, as described in \autoref{sec:data}. The mode will receive an example $\boldsymbol{x}_i$ , and, if wrong, will then perform a corrective update. This will be done for all 58 hard FPs in random order. This process will be repeated 200 times with a different random ordering each time, to get a distributional understanding of the true impact. This avoids biasing our results to an ordering that may be particular (un)favorable to any approach. We will look at the global metrics and how the change as more hard FPs are presented, and the total number of errors/times the model had to be adjusted against hard FPs. Ideally, only 1 error/adjustment would be needed for the model to get all remaining hard FPs correct.

\subsection{Baseline Performance}

Now that we have specified our approach to this problem and the baseline methods we will compare against; we will first look at the performance on the larger test set of 276,921 macro files.  Our goal is to determine the classifier’s performance against the global population at-large. This can be found in \autoref{tbl:stnd_results}. Algorithms prepended with "MalConv" work base of first training the MalConv architecture. All others used the production domain knowledge-based feature extractor.

\begin{table}[!htb]
\caption{Performance of all 4 methods on the larger test set. } \label{tbl:stnd_results}
\adjustbox{max width=\columnwidth}{%
\begin{tabular}{lccccc}
\toprule
\multicolumn{1}{c}{Algorithm} & Acc   & AUC   & $\text{AUC}_{FPR\leq.1\%}$ & FPR         & TPR         \\ \midrule
MalConv+PA                    & 96.66 & 99.34 & 78.30                  & 0.1005      & 58.35    \\
MalConv+SGD                   & 97.06 & 99.36 & 79.21                  & 0.0997      & 66.18    \\
MalConv+Prototype             & 60.97 & 64.96 & 50.01                  & 13.29       & 86.70    \\
GBDT                          & 99.85 & 99.97 & 99.27                  & 0.0930      & 99.65    \\ 
PA                            & 95.13 & 97.12 & 50.39                  & 0.1006      & 2.310    \\
Kernel PA                     & 66.80 & 63.26 & 56.28                  & 0.0999      & 14.87    \\
\bottomrule
\end{tabular}
}
\end{table}

Prototype-Networks, a few-shot learning approach, is most notable for its deficient performance and lack of generalization to the test set. While the few-shot paradigm framework is an attractive model to use for the issue of fixing false positives at deployment time, we found significant problems in training and using in --- often resulting in network divergence during training, or the most successful case was networks that degraded to degenerate solutions of projecting all data onto a few sets of points. This caused a limited number of distance values to each class centroid, making it difficult to achieve the desired FPR and hampered overall performance. 

We believe this stems from the miss-match in the underlying assumptions used in the few-shot learning literature and our application space. For most few-shot datasets, e.g., omniglot, there are many classes with a limited amount of intraclass variance. This makes the problem of recognizing a new class and encapsulating it easier. For our data, we have only two classes, each of which has significant intraclass variance. In particular, benign applications have broad diversity in their content and use cases. This clashes with the few-shot models’ underlying assumptions, like Prototype-networks, leading to the issues we see.  

We also see that the GBDT model using the domain knowledge features has the absolute best performance by a wide margin. While the PA and SGD approaches that use MalConv as a feature extractor have close AUC scores, the tail of the distribution we care about is an FPR $\approx$ 0.1\%. In this domain the GBDT, using domain knowledge features, clearly has the best performance with an up to 70.8\% improvement in TPR at the desired threshold. MalConv+PA and MalConv+SGD have performance similar to each other, with detection rates considered acceptable but not as good as GBDT. However, the MalConv+SGD combination appears to perform better initially.

We emphasize that while the GBDT model has much better \textit{global} TPR, these values are not useful if the \textit{local} FPR is too high. In such a situation, it is preferable to switch to a model with lower but acceptable TPR to provide some level of defense because the alternative is a user disabling the tools, leaving them vulnerable. This is the precise situation that has occurred in production that our work mitigates. We show this in the next section.

\subsection{Hard Benign Sample Performance}

Given these initial results, our preference would lean towards using domain knowledge features in a GBDT model. This approach would allow us to provide users with some interpretability to reduce the "black-box" feel of malware classification. However, the picture changes when we look at the population of 58 samples generated by popular enterprise software. These results are shown in \autoref{tbl:hard_results}. Using a fixed global model, the GBDT fails to classify all of these critical applications. This necessitates using a allowlist to suppress alerts on these files. Because these are document, the hash representation changes with each update/save operation resulting in regular and persistent false positives that have lead to significant user frustration.

\begin{table}[!htb]
\centering
\caption{False Positive Rate (FPR) on the 58 hard held out samples, lower is better. The \textit{Fixed} column shows the FPR with a static model. The Adaptive column shows FPR as the model gets a chance to update after each error, sampled over 200 trials.  } \label{tbl:hard_results}
\adjustbox{width=0.8\columnwidth}{%
\begin{tabular}{@{}lcc@{}}
\toprule
                              & \multicolumn{2}{c}{Hard FP Rate (\%)} \\ \cmidrule(l){2-3} 
\multicolumn{1}{c}{Algorithm} & Fixed       & Adaptive           \\ \midrule
MalConv+PA                    & 58.62\%       & \textbf{\phantom{0}4.33$\pm$1.919}\%       \\
MalConv+SGD                   & 37.93\%       & 26.46$\pm$1.893\%       \\
GBDT                          & 100.0\%       & N/A                \\ \bottomrule
\end{tabular}
}
\end{table}

Both PA and SGD based approaches can run in the adaptive scenario. Again, in this situation, the 58 hard benign files are presented sequentially as if being found over time in deployment. When the model makes an FP error (calling the benign application malicious), we perform a model update to hopefully resolve the issue. This makes the test order sensitive, so we averaged the results over 200 random trials. 

The PA approach, while having a 58.6\% FPR under the fixed scenario, drops down to just 4.33\% in the adaptive scenario. This corresponds to making an average of \textit{just 2.51 errors} against all 58 hard benign applications. This is better than the GBDT, which can not adapt on a single-instance basis, and better than the SGD method. SGD, while it does reduce its FPR from 37.9\% down to 26.5\%, makes smaller changes in a model view that it is trying to optimize a larger function. This shows how the PA model's attempt to fully correct each error better matches this issue in production and leads to efficiency in fixing FPs with a minimal number of errors.

\begin{figure}[!h]
    \centering
    \adjustbox{max width=\columnwidth}{%
\begin{tikzpicture}

\definecolor{color0}{rgb}{0.12156862745098,0.466666666666667,0.705882352941177}
\definecolor{color1}{rgb}{1,0.498039215686275,0.0549019607843137}

\begin{axis}[
tick align=outside,
tick pos=left,
x grid style={white!69.0196078431373!black},
xlabel={Method},
xmin=-0.5, xmax=1.5,
xtick style={color=black},
xtick={0,1},
xticklabels={MalConv+SGD,MalConv+PA},
y grid style={white!69.0196078431373!black},
ylabel={Errors},
ymin=0.0827221411564956, ymax=18.8652486285493,
ytick style={color=black}
]
\addplot [only marks, mark=*, draw=white!24.7058823529412!black, fill=color0, colormap/viridis]
table{%
x                      y
-1.11022302462516e-16 13
-0.0313620071684588 13
0.0313620071684587 13
-0.0627240143369177 13
0.0627240143369174 13
-0.0940860215053764 13
0.0940860215053763 13
-1.11022302462516e-16 14
-0.0313620071684588 14
0.0313620071684587 14
-0.0627240143369177 14
0.0627240143369174 14
-0.0940860215053764 14
0.0940860215053763 14
-0.125448028673835 14
0.125448028673835 14
-0.156810035842294 14
0.156810035842294 14
-0.188172043010753 14
0.188172043010753 14
-0.219534050179212 14
0.219534050179211 14
-0.25089605734767 14
0.25089605734767 14
-0.282258064516129 14
0.282258064516129 14
-0.313620071684588 14
0.313620071684588 14
-0.344982078853047 14
0.344982078853046 14
-0.376344086021505 14
0.376344086021505 14
-0.4 14
0.4 14
-0.4 14
0.4 14
-0.4 14
0.4 14
-0.4 14
0.4 14
-0.4 14
0.4 14
-0.4 14
-1.11022302462516e-16 15
-0.0313620071684588 15
0.0313620071684587 15
-0.0627240143369177 15
0.0627240143369174 15
-0.0940860215053764 15
0.0940860215053763 15
-0.125448028673835 15
0.125448028673835 15
-0.156810035842294 15
0.156810035842294 15
-0.188172043010753 15
0.188172043010753 15
-0.219534050179212 15
0.219534050179211 15
-0.25089605734767 15
0.25089605734767 15
-0.282258064516129 15
0.282258064516129 15
-0.313620071684588 15
0.313620071684588 15
-0.344982078853047 15
0.344982078853046 15
-0.376344086021505 15
0.376344086021505 15
-0.4 15
0.4 15
-0.4 15
0.4 15
-0.4 15
0.4 15
-0.4 15
0.4 15
-0.4 15
0.4 15
-0.4 15
0.4 15
-0.4 15
0.4 15
-0.4 15
0.4 15
-0.4 15
0.4 15
-0.4 15
0.4 15
-0.4 15
0.4 15
-0.4 15
0.4 15
-0.4 15
0.4 15
-0.4 15
0.4 15
-0.4 15
0.4 15
-0.4 15
0.4 15
-0.4 15
0.4 15
-0.4 15
0.4 15
-0.4 15
0.4 15
-0.4 15
0.4 15
-0.4 15
0.4 15
-0.4 15
0.4 15
-0.4 15
0.4 15
-0.4 15
-1.11022302462516e-16 16
-0.0313620071684588 16
0.0313620071684587 16
-0.0627240143369177 16
0.0627240143369174 16
-0.0940860215053764 16
0.0940860215053763 16
-0.125448028673835 16
0.125448028673835 16
-0.156810035842294 16
0.156810035842294 16
-0.188172043010753 16
0.188172043010753 16
-0.219534050179212 16
0.219534050179211 16
-0.25089605734767 16
0.25089605734767 16
-0.282258064516129 16
0.282258064516129 16
-0.313620071684588 16
0.313620071684588 16
-0.344982078853047 16
0.344982078853046 16
-0.376344086021505 16
0.376344086021505 16
-0.4 16
0.4 16
-0.4 16
0.4 16
-0.4 16
0.4 16
-0.4 16
0.4 16
-0.4 16
0.4 16
-0.4 16
0.4 16
-0.4 16
0.4 16
-0.4 16
0.4 16
-0.4 16
0.4 16
-0.4 16
0.4 16
-0.4 16
0.4 16
-0.4 16
0.4 16
-0.4 16
0.4 16
-0.4 16
0.4 16
-0.4 16
0.4 16
-0.4 16
0.4 16
-0.4 16
-1.11022302462516e-16 17
-0.0313620071684588 17
0.0313620071684587 17
-0.0627240143369177 17
0.0627240143369174 17
-0.0940860215053764 17
0.0940860215053763 17
-0.125448028673835 17
0.125448028673835 17
-0.156810035842294 17
0.156810035842294 17
-0.188172043010753 17
0.188172043010753 17
-0.219534050179212 17
0.219534050179211 17
-0.25089605734767 17
0.25089605734767 17
-0.282258064516129 17
0.282258064516129 17
-0.313620071684588 17
-1.11022302462516e-16 18
-0.0313620071684588 18
0.0313620071684587 18
-0.0627240143369177 18
0.0627240143369174 18
-0.0940860215053764 18
0.0940860215053763 18
};
\addplot [only marks, mark=*, draw=white!24.7058823529412!black, fill=color1, colormap/viridis]
table{%
x                      y
1 1
0.968637992831541 1
1.03136200716846 1
0.937275985663083 1
1.06272401433692 1
0.905913978494624 1
1.09408602150538 1
0.874551971326165 1
1.12544802867384 1
0.843189964157706 1
1.15681003584229 1
0.811827956989247 1
1.18817204301075 1
0.780465949820789 1
1.21953405017921 1
0.74910394265233 1
1.25089605734767 1
0.717741935483871 1
1.28225806451613 1
0.686379928315412 1
1.31362007168459 1
0.655017921146954 1
1.34498207885305 1
0.623655913978495 1
1.37634408602151 1
0.6 1
1.4 1
0.6 1
1.4 1
0.6 1
1.4 1
0.6 1
1.4 1
0.6 1
1.4 1
0.6 1
1.4 1
0.6 1
1 2
0.968637992831541 2
1.03136200716846 2
0.937275985663083 2
1.06272401433692 2
0.905913978494624 2
1.09408602150538 2
0.874551971326165 2
1.12544802867384 2
0.843189964157706 2
1.15681003584229 2
0.811827956989247 2
1.18817204301075 2
0.780465949820789 2
1.21953405017921 2
0.74910394265233 2
1.25089605734767 2
0.717741935483871 2
1.28225806451613 2
0.686379928315412 2
1.31362007168459 2
0.655017921146954 2
1.34498207885305 2
0.623655913978495 2
1.37634408602151 2
0.6 2
1.4 2
0.6 2
1.4 2
0.6 2
1.4 2
0.6 2
1.4 2
0.6 2
1.4 2
0.6 2
1.4 2
0.6 2
1.4 2
0.6 2
1.4 2
0.6 2
1.4 2
0.6 2
1.4 2
0.6 2
1.4 2
0.6 2
1.4 2
0.6 2
1.4 2
0.6 2
1.4 2
0.6 2
1.4 2
0.6 2
1.4 2
0.6 2
1.4 2
0.6 2
1.4 2
0.6 2
1.4 2
0.6 2
1.4 2
0.6 2
1.4 2
0.6 2
1.4 2
0.6 2
1.4 2
1 3
0.968637992831541 3
1.03136200716846 3
0.937275985663083 3
1.06272401433692 3
0.905913978494624 3
1.09408602150538 3
0.874551971326165 3
1.12544802867384 3
0.843189964157706 3
1.15681003584229 3
0.811827956989247 3
1.18817204301075 3
0.780465949820789 3
1.21953405017921 3
0.74910394265233 3
1.25089605734767 3
0.717741935483871 3
1.28225806451613 3
0.686379928315412 3
1.31362007168459 3
0.655017921146954 3
1.34498207885305 3
0.623655913978495 3
1.37634408602151 3
0.6 3
1.4 3
0.6 3
1.4 3
0.6 3
1.4 3
0.6 3
1.4 3
0.6 3
1.4 3
0.6 3
1.4 3
0.6 3
1.4 3
0.6 3
1.4 3
0.6 3
1.4 3
0.6 3
1.4 3
0.6 3
1.4 3
0.6 3
1.4 3
0.6 3
1.4 3
0.6 3
1.4 3
0.6 3
1 4
0.968637992831541 4
1.03136200716846 4
0.937275985663083 4
1.06272401433692 4
0.905913978494624 4
1.09408602150538 4
0.874551971326165 4
1.12544802867384 4
0.843189964157706 4
1.15681003584229 4
0.811827956989247 4
1.18817204301075 4
0.780465949820789 4
1.21953405017921 4
0.74910394265233 4
1.25089605734767 4
0.717741935483871 4
1.28225806451613 4
0.686379928315412 4
1.31362007168459 4
0.655017921146954 4
1.34498207885305 4
0.623655913978495 4
1.37634408602151 4
1 5
0.968637992831541 5
1.03136200716846 5
0.937275985663083 5
1.06272401433692 5
0.905913978494624 5
1.09408602150538 5
0.874551971326165 5
1.12544802867384 5
0.843189964157706 5
1.15681003584229 5
0.811827956989247 5
};
\end{axis}

\end{tikzpicture}
    }
    \caption{Number of errors made in the adaptive scenario. Each dot representing the result from one trial with a different random ordering of the benign files. }
    \label{fig:adaptive_errors}
\end{figure}
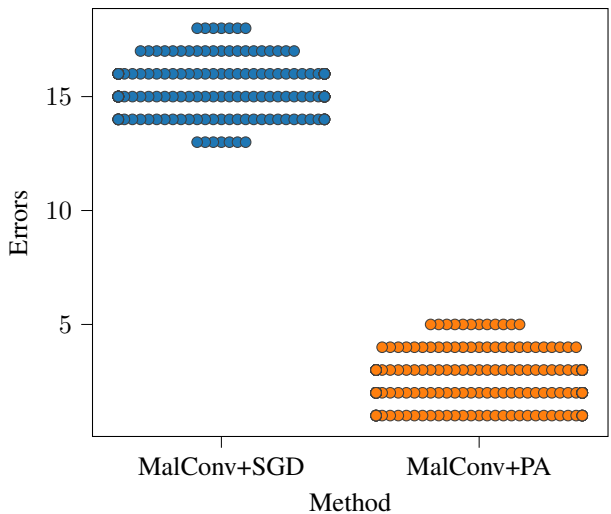

In particular, \autoref{fig:adaptive_errors} shows results from all 200 random trials for each method. The MalConv+PA approach never makes more than 5 errors on all 58 files, and as few as just 1 error. The MalConv+SGD approach makes \textit{at least} 13 errors and up to 18, making it several times worse on this critical dataset.

While the MalConv+PA model did not have the highest possible accuracy on the global dataset, customers’ ability to adapt the model to their local domain is critical. This new approach to using the PA model is the only current method to quickly adapt deployed models, making it preferable to deal with such hard benign programs. We also note that the change in TPR and FPR trade-offs can be fixed easily by sending the PA weights (a few KB) back to the AV provider for re-calibration of the threshold to obtain an FPR of 0.1\% again, which restores TPR to 58\%, and still classifiers all hard FPs correctly.

\subsection{Adaptive Impact and Estimated Impact }

Now that we have shown our PA approach is the best current method to deal with false positives in production, we consider the global impact of these adjustments on the MalConv+PA model, as well as our ability to estimate the impact described using our k-means based approach from \autoref{sec:kmeans_impact}. We arbitrarily choose $K=1024$ clusters based on two factors. Any $K$ smaller than the whole dataset accomplishes the privacy goal, and larger values of $K$ are expected to produce better AUC approximations, as the limit is $K$ equal to the size of the corpus, resulting in the true AUC being computed. 

\begin{figure}
    \centering
    \adjustbox{max width=\columnwidth}{%
        \input{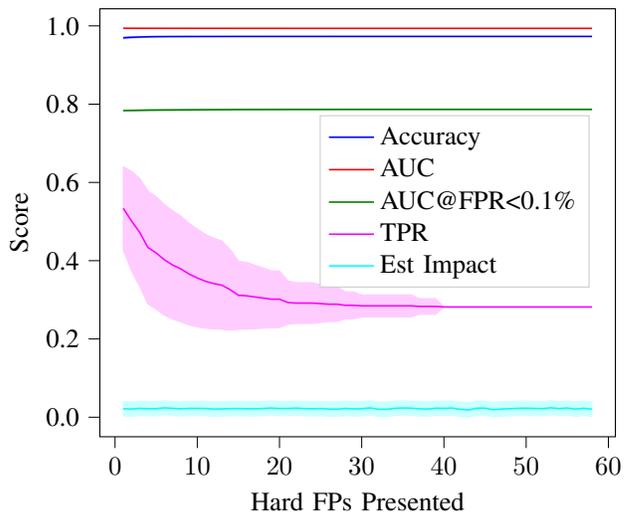}
    }
    \caption{Accuracy, AUC, $\text{AUC}_{FPR}\leq 0.1\%$, TPR, and estimated impact on the Macro+PA model larger test dataset as hard benign files are presented. Average results over 200 random trials, standard deviation shown in lighter bands around each method. The first three methods, standard deviation is drawn but exceptionally small (zoom in digitally to see). }
    \label{fig:macro_active_errs}
\end{figure}

We start by looking at all of our metrics, excluding the FPR, in \autoref{fig:macro_active_errs}. These results are over the average of 200 trials and measured against the larger test set of 276,921 files. Here we can see that the global ability of the model is not significantly altered. Using the standard threshold of 0, the Accuracy and AUC see minimal impacts, and on average, actually improve slightly as more files are presented. For example, Accuracy has the biggest improvement from 96.95\% to 97.30\%.  

The estimated impact, using \autoref{algo:estimate_impact}, of the model is always measured against the original global model. This is done so that we are not lured into a false sense of confidence, and we want to know if the model has diverged too significantly from the global case. If we measured impact based on changes against previous versions of the model, the impact would almost always be 0 because the PA approach makes $\leq 5$ changes over all 58 files. Measuring against the previously modified model could also lead to many small changes with collectively represent a large change from the global model, without realizing that such significant changes have occurred. 

Measuring against the original global model, \autoref{fig:macro_active_errs} shows that the estimated impact to AUC is always small in magnitude and usually positive. The actual AUC (the red line) shows minimal impact and a slight positive trend. This indicates a behavioral match between our estimation approach and the test data appears to be consistent. We also note that the AUC measured up to a maximum FPR of 0.1\% also shows no significant impact, with only slight improvement.

We note that measuring the AUC up to a maximum FPR involves determining the exact threshold for achieving that FPR, requiring all the data. So these results show that the global properties of our model are maintained, but results could differ based on the threshold originally learned from the training data, which may no longer be the best threshold to use in practice. The True Positive Rate (purple) in \autoref{fig:macro_active_errs} shows this result when using the original threshold, which reduces the TPR by $48.3\%$. This is not surprising because we are explicitly biasing the model toward calling more things benign, but still effective at 28.17 TPR.

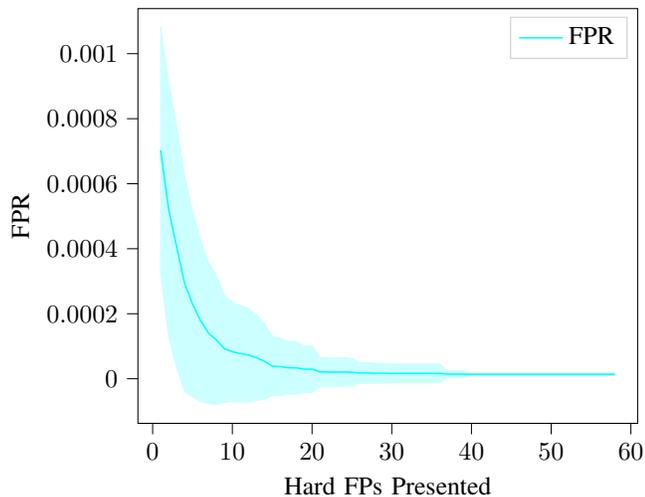
\begin{figure}[!h]
    \centering
    \adjustbox{max width=\columnwidth}{%
\begin{tikzpicture}

\definecolor{color0}{rgb}{0,1,1}

\begin{axis}[
legend cell align={left},
legend style={fill opacity=0.8, draw opacity=1, text opacity=1, draw=white!80!black},
tick align=outside,
tick pos=left,
x grid style={white!69.0196078431373!black},
xlabel={Hard FPs Presented},
xmin=-1.85, xmax=60.85,
xtick style={color=black},
y grid style={white!69.0196078431373!black},
ylabel={FPR},
ymin=-0.00013715456677628, ymax=0.0011389688584879,
ytick style={color=black},
yticklabel style={
        /pgf/number format/fixed,
        /pgf/number format/precision=5
},
scaled y ticks=false
]
\path [draw=color0, fill=color0, opacity=0.2]
(axis cs:1,0.00108096324824862)
--(axis cs:1,0.000324444230178992)
--(axis cs:2,0.000130823500468256)
--(axis cs:3,3.48148177525377e-05)
--(axis cs:4,-3.80524444550641e-05)
--(axis cs:5,-5.56518469516796e-05)
--(axis cs:6,-6.93292246295031e-05)
--(axis cs:7,-7.71892971005438e-05)
--(axis cs:8,-7.91489565369993e-05)
--(axis cs:9,-7.23668957354075e-05)
--(axis cs:10,-6.90495921937742e-05)
--(axis cs:11,-7.08849819568501e-05)
--(axis cs:12,-7.09810028915212e-05)
--(axis cs:13,-6.65740010679782e-05)
--(axis cs:14,-6.36716998654492e-05)
--(axis cs:15,-5.2109994307322e-05)
--(axis cs:16,-5.05111312465043e-05)
--(axis cs:17,-4.6880363146294e-05)
--(axis cs:18,-4.72177631678151e-05)
--(axis cs:19,-4.19495488592071e-05)
--(axis cs:20,-4.19495488592071e-05)
--(axis cs:21,-2.32130227667742e-05)
--(axis cs:22,-2.31638158146753e-05)
--(axis cs:23,-2.31638158146753e-05)
--(axis cs:24,-2.31638158146753e-05)
--(axis cs:25,-2.25552983389293e-05)
--(axis cs:26,-1.40229084534559e-05)
--(axis cs:27,-1.41239786660019e-05)
--(axis cs:28,-1.35815217243112e-05)
--(axis cs:29,-1.35815217243112e-05)
--(axis cs:30,-1.26992052379454e-05)
--(axis cs:31,-1.27438276696508e-05)
--(axis cs:32,-1.27438276696508e-05)
--(axis cs:33,-1.27438276696508e-05)
--(axis cs:34,-1.27438276696508e-05)
--(axis cs:35,-1.27438276696508e-05)
--(axis cs:36,-1.27438276696508e-05)
--(axis cs:37,4.13857034537266e-06)
--(axis cs:38,4.13857034537266e-06)
--(axis cs:39,4.13857034537266e-06)
--(axis cs:40,9.58266564558494e-06)
--(axis cs:41,9.58266564558494e-06)
--(axis cs:42,9.58266564558494e-06)
--(axis cs:43,9.58266564558494e-06)
--(axis cs:44,9.58266564558494e-06)
--(axis cs:45,9.58266564558494e-06)
--(axis cs:46,9.58266564558494e-06)
--(axis cs:47,9.58266564558494e-06)
--(axis cs:48,9.58266564558494e-06)
--(axis cs:49,9.58266564558494e-06)
--(axis cs:50,9.58266564558494e-06)
--(axis cs:51,9.58266564558494e-06)
--(axis cs:52,9.58266564558494e-06)
--(axis cs:53,9.58266564558494e-06)
--(axis cs:54,9.58266564558494e-06)
--(axis cs:55,9.58266564558494e-06)
--(axis cs:56,9.58266564558494e-06)
--(axis cs:57,9.58266564558494e-06)
--(axis cs:58,9.58266564558494e-06)
--(axis cs:58,1.77231828683173e-05)
--(axis cs:58,1.77231828683173e-05)
--(axis cs:57,1.77231828683173e-05)
--(axis cs:56,1.77231828683173e-05)
--(axis cs:55,1.77231828683173e-05)
--(axis cs:54,1.77231828683173e-05)
--(axis cs:53,1.77231828683173e-05)
--(axis cs:52,1.77231828683173e-05)
--(axis cs:51,1.77231828683173e-05)
--(axis cs:50,1.77231828683173e-05)
--(axis cs:49,1.77231828683173e-05)
--(axis cs:48,1.77231828683173e-05)
--(axis cs:47,1.77231828683173e-05)
--(axis cs:46,1.77231828683173e-05)
--(axis cs:45,1.77231828683173e-05)
--(axis cs:44,1.77231828683173e-05)
--(axis cs:43,1.77231828683173e-05)
--(axis cs:42,1.77231828683173e-05)
--(axis cs:41,1.77231828683173e-05)
--(axis cs:40,1.77231828683173e-05)
--(axis cs:39,2.44712091368901e-05)
--(axis cs:38,2.44712091368901e-05)
--(axis cs:37,2.44712091368901e-05)
--(axis cs:36,4.51886982353268e-05)
--(axis cs:35,4.51886982353268e-05)
--(axis cs:34,4.51886982353268e-05)
--(axis cs:33,4.51886982353268e-05)
--(axis cs:32,4.51886982353268e-05)
--(axis cs:31,4.51886982353268e-05)
--(axis cs:30,4.52207776252896e-05)
--(axis cs:29,4.74837269016843e-05)
--(axis cs:28,4.74837269016843e-05)
--(axis cs:27,4.96369220984084e-05)
--(axis cs:26,4.9689255529199e-05)
--(axis cs:25,6.22101401414221e-05)
--(axis cs:24,6.44293958722016e-05)
--(axis cs:23,6.44293958722016e-05)
--(axis cs:22,6.44293958722016e-05)
--(axis cs:21,6.55524283276562e-05)
--(axis cs:20,0.000101240057008776)
--(axis cs:19,0.000101240057008776)
--(axis cs:18,0.00011417845348421)
--(axis cs:17,0.00011468477350104)
--(axis cs:16,0.000124605090978048)
--(axis cs:15,0.00012996234330061)
--(axis cs:14,0.000171974672061039)
--(axis cs:13,0.000197043799725696)
--(axis cs:12,0.000217404780456238)
--(axis cs:11,0.000225669258083408)
--(axis cs:10,0.000235569247035576)
--(axis cs:9,0.000256911478669252)
--(axis cs:8,0.000317691621925302)
--(axis cs:7,0.000358608280801407)
--(axis cs:6,0.000427296626355294)
--(axis cs:5,0.00051532586420959)
--(axis cs:4,0.000624974783860626)
--(axis cs:3,0.000778761404682746)
--(axis cs:2,0.000913241696080163)
--(axis cs:1,0.00108096324824862)
--cycle;

\addplot [semithick, color0]
table {%
1 0.000702703739213807
2 0.00052203259827421
3 0.000406788111217642
4 0.000293461169702781
5 0.000229837008628955
6 0.000178983700862896
7 0.000140709491850431
8 0.000119271332694151
9 9.22722914669222e-05
10 8.32598274209011e-05
11 7.73921380632789e-05
12 7.32118887823585e-05
13 6.5234899328859e-05
14 5.41514860977947e-05
15 3.89261744966442e-05
16 3.70469798657717e-05
17 3.39022051773729e-05
18 3.34803451581975e-05
19 2.96452540747842e-05
20 2.96452540747842e-05
21 2.1169702780441e-05
22 2.06327900287632e-05
23 2.06327900287632e-05
24 2.06327900287632e-05
25 1.98274209012464e-05
26 1.78331735378715e-05
27 1.77564717162033e-05
28 1.69511025886865e-05
29 1.69511025886865e-05
30 1.62607861936721e-05
31 1.6222435282838e-05
32 1.6222435282838e-05
33 1.6222435282838e-05
34 1.6222435282838e-05
35 1.6222435282838e-05
36 1.6222435282838e-05
37 1.43048897411314e-05
38 1.43048897411314e-05
39 1.43048897411314e-05
40 1.36529242569511e-05
41 1.36529242569511e-05
42 1.36529242569511e-05
43 1.36529242569511e-05
44 1.36529242569511e-05
45 1.36529242569511e-05
46 1.36529242569511e-05
47 1.36529242569511e-05
48 1.36529242569511e-05
49 1.36529242569511e-05
50 1.36529242569511e-05
51 1.36529242569511e-05
52 1.36529242569511e-05
53 1.36529242569511e-05
54 1.36529242569511e-05
55 1.36529242569511e-05
56 1.36529242569511e-05
57 1.36529242569511e-05
58 1.36529242569511e-05
};
\addlegendentry{FPR}
\end{axis}

\end{tikzpicture}
    }
    \caption{False Positive Rate (y-axis) as the average number of hard FPs (x-axis) are presented to the MalConv+PA model. Because we do not alter the threshold used, and the hard FPs are corrected in favor of calling more things benign, the FPR goes down significantly by $74\times$.}
    \label{fig:macro_active_fpr}
\end{figure}

\begin{figure}[!h]
    \centering
    \adjustbox{max width=\columnwidth}{%
        \input{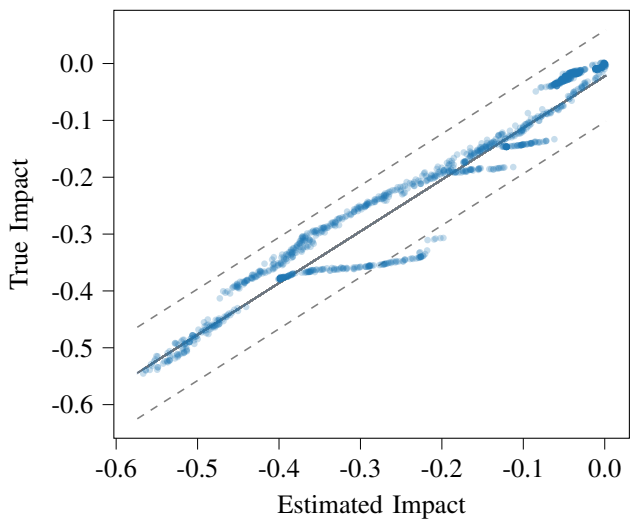}
    }
    \caption{Estimated Impact to AUC (x-axis) compared to actual impact (y-axis). Black line shows a simple linear fit, and dashed lines show the 95\% confidence interval of the results. Results shown for a random sample of 2,500 points for legibility. }
    \label{fig:est_impact}
\end{figure}

\begin{figure*}
    \centering
    \adjustbox{max width=\textwidth}{%
        \input{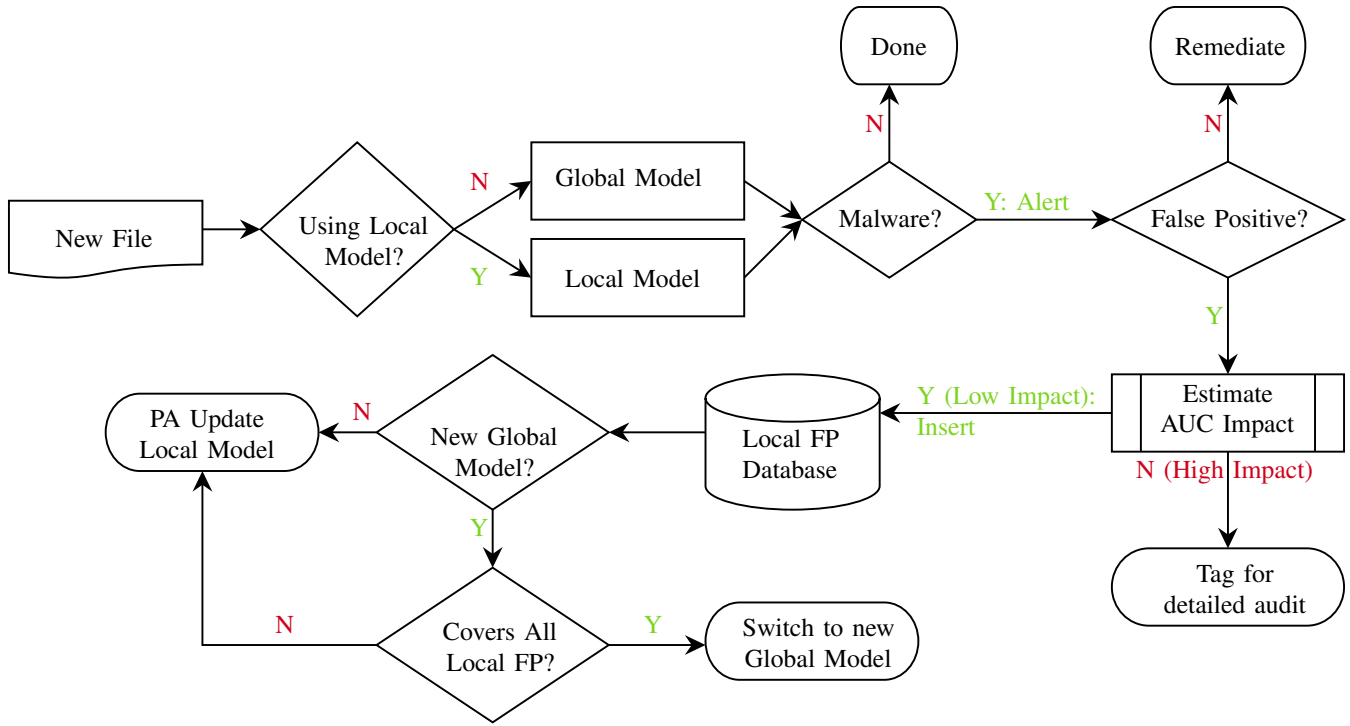}
    }
    \caption{Flow diagram for deployed use. A new file comes in and must be classified, by default using the global model (GBDT). Alerts are reviewed and if determined to be a false positive, the estimated impact is computed with \autoref{algo:estimate_impact}. If above some threshold, the file is tagged for expert audit. Otherwise the model can be adaptive with low impact to performance. The datum is added to a local database. If a new global model is available and satisfies all local historical FPs, switch to the new global model. Otherwise continue using local model, and perform a Passive Aggressive update to correct local errors, and switch to using local model.   }
    \label{fig:flow}
\end{figure*}

The drop in TPR comes with the benefit of a far more significant drop in the FPR of the models. This is shown in \autoref{fig:macro_active_fpr}, where the TPR decreases monotonically down to 0.001365\%, a $73\times$ reduction in false positives. We note that re-calibrating the threshold to achieve a 0.1\% false positive rate still results in 0 FPs on the hard benign applications, since the PA approach guarantees that they are on the malicious side of the margin, and do not need any special thresholding to be classified correctly. We will talk about re-calibration more in \autoref{sec:practical}.

These results show that the global properties of the MalConv+PA model have remained adequate and unaltered by our approach, baring a potential re-calibration of the threshold to achieve an FPR of 0.1\% instead of 0.001\%. While our testing showed that our AUC impact estimation from \autoref{sec:kmeans_impact} correctly predicted low impact to the model, it does not tell us if our approach would successfully stop us from making a dangerously large change to the model.

To test that, we ran an experiment where we took files from the training set, swapped their labels, and then performed PA updates. We then compared the estimated impact from \autoref{algo:estimate_impact} with the actual impact on the larger test set. These results can be found in \autoref{fig:est_impact}. Here we can see that our approach reliably estimates the true impact to AUC in a strong linear relationship, with almost all points fitting with a 95\% confidence interval. Thus we can rely on our approach to detect large changes to AUC (e.g., -0.5 would indicate taking the model from accurate to near random-guessing levels of performance) that would be detrimental to the model's use. A threshold of maximal AUC change can be selected such that we either warn the users that the change may reduce their ability to detect malware reliably or, if large enough, prevent the users from making the change entirely --- and potentially warning that the file under consideration needs expert review.

\section{Practical Considerations} \label{sec:practical}

Now that we have described our approach to adapting a production model based on false positives, we take a moment to consider how this system can be used in production. The first concern is that, under the global model, the GBDT approach does have better performance than MalConv+PA. For this reason, we would deploy our approach using the GBDT based model first and only switch to the MalConv+PA model once a hard FP has been presented that needs to be corrected. This is because not all customers experience these hard FPs, and we can retain the better global benefits of GBDT in such a case. 

Once a hard FP has been correct, and we have switched to MalConv+PA as the primary model, we would still encourage users to send the false positives back for further analysis and enter the process for a global model update. A future GBDT version, given many such FPs, may eventually get all hard FPs for a client, and we can switch back to this better global model. The MalConv+PA approach gives us the means to make sure those customers have an immediate and still useful solution, without causing them to forgo all protection due to excessive alerts and multi-week to multi-month lag time of the global model. A flow diagram of this combined process is given in \autoref{fig:flow}. 

In such situations, we may also choose to run the GBDT model and MalConv+PA model simultaneously on all inputs, while relying on MalConv+PA to act as an \textit{abstainer} for primary alerts. For cases where the two models disagree, we may throw lower-priority alerts or request that such samples be sent back for further analysis and eventually improve the global model. 

Finally, we note that the current approach results in an effective change in the target FPR rate after MalConv+PA performs a correction. Rather than let the customer run at their model at this new effective FPR, we can send the corrected \textit{model} back to the corporate environment to perform a re-calibration on the threshold. This is computationally cheap to perform, requiring only a few seconds on our corpus, and eliminates the reduction in TPR observed in \autoref{fig:macro_active_errs}. Sending the model back to corporate and returning a new threshold is done because we do not want to distribute gigabytes of training data to customers as part of a deployed model. This would pose an unnecessary storage cost and privacy concerns. This re-calibration can be done by sending only the changed model back to corporate, protecting the customer's privacy of samples if they are unwilling or unable to share them for global model improvement.

\section{Conclusion} \label{sec:conclusion}

Using machine learning to detect malicious activity will always yield false positives in production environments. That is the trade-off between detections that only catch the \emph{known} versus detections that also attempt to catch the \emph{unknown}. However, the current false positive triage approaches are antiquated, often relying solely on “whack-a-mole” hash-based allowlisting that leaves security workers drowning in alerts. There is an opportunity to develop novel methods to significantly reduce local false positives by exposing intuitive ML capabilities to security workers, so their domain expertise can refine and tailor models.

Our approach highlights how an iterative, human-in-the-loop process can be applied to close the gap between identifying false positives local to an environment and ensuring future models do not repeat those mistakes. We aim to limit the set of observations required to tailor a model while keeping an eye on preserving user data privacy. 

\renewcommand*{\bibfont}{\small}

{\small 
\printbibliography 
}

\end{document}